\pdfoutput=1

\documentclass[11pt]{article}
\usepackage{tcolorbox}
\tcbuselibrary{breakable}
\usepackage{CJKutf8}

\definecolor{PromptFrame}{HTML}{65758B}
\definecolor{PromptBack}{HTML}{F7F9FC}
\definecolor{PromptTitleBack}{HTML}{E7EEF7}
\definecolor{PromptAccent}{HTML}{355C8A}

\newtcolorbox{promptbox}[1]{
  breakable,
  colback=PromptBack,
  colframe=PromptFrame,
  colbacktitle=PromptTitleBack,
  coltitle=black,
  fonttitle=\bfseries\small,
  fontupper=\small,
  title={#1},
  title after break={#1 (continued)},
  boxrule=0.55pt,
  arc=1.2mm,
  left=1.5mm,
  right=1.5mm,
  top=1.2mm,
  bottom=1.2mm,
  before skip=7pt,
  after skip=8pt
}

\newcommand{\promptrole}[1]{%
  \textbf{\textcolor{PromptAccent}{#1}}%
}

\newcommand{\promptvar}[1]{%
  \texttt{\textcolor{PromptAccent}{[#1]}}%
}

\newcommand{\promptdivider}{%
  \par\medskip
  {\color{PromptFrame}\hrule height 0.4pt}
  \medskip
}

\definecolor{primary}{RGB}{253,248,225}
\definecolor{bandgray}{RGB}{240,240,240}
\newcommand{\hl}{\cellcolor{primary}}

\usepackage[preprint]{acl}
\usepackage{placeins}
\usepackage{times}
\usepackage{latexsym}
\usepackage{booktabs}
\usepackage{placeins}
\usepackage[T1]{fontenc}
\usepackage[utf8]{inputenc}
\usepackage{multirow}
\usepackage{microtype}
\usepackage{inconsolata}
\usepackage{xcolor}
\usepackage{colortbl}

\usepackage{amsmath}
\usepackage{amssymb}

\usepackage{booktabs}
\usepackage{graphicx}
\usepackage{url}
\usepackage{enumitem}

\title{MANTIS: Multi-Axis News Transformations for Instance-Wise Stress Testing and Worst-Case Training in Fake News Detection}

\author{
\textbf{Dexin Liu}$^{1,2,\star}$ \quad
\textbf{Fan Yang}$^{3,\star}$ \quad
\textbf{Zimu Wang}$^{1,2,\star}$ \quad
\textbf{Xunfei Zhu}$^{3}$ \\
\textbf{Bolin Ren}$^{1}$ \quad
\textbf{Bingyan Zhang}$^{3}$ \quad
\textbf{Wei Wang}$^{1}$ \quad
\textbf{Qi Chen}$^{1,\dagger}$ \\
{\normalfont\small $^{1}$Xi'an Jiaotong-Liverpool University, Suzhou, China} \\
{\normalfont\small $^{2}$University of Liverpool, Liverpool, United Kingdom} \\
{\normalfont\small $^{3}$Nanjing Sibei Intelligent Technology Co., Ltd., Nanjing, China} \\
{\normalfont\small \texttt{Dexin.Liu24@student.xjtlu.edu.cn} \quad
\texttt{Qi.Chen02@xjtlu.edu.cn}} \\
{\normalfont\small $^{\star}$Equal contribution. $^{\dagger}$Corresponding author.}
}
\begin{document}
\maketitle

\begin{abstract} 

Automated fake news detection is vital on social media, where fake news is easily rewritten to fool detectors. However, existing approaches typically rely on superficial writing cues or single-style text rewrites, resulting in poor robustness under diverse cosmetic modifications. We construct \textbf{MANTIS-Bench}, a multi-axis presentation-attack benchmark that pairs original news with rewritten views across writing style, emotional expression, and dissemination framing. We further propose \textbf{MANTIS}, an instance-adaptive adversarial framework enabling dynamic worst-case training tailored to each news article with zero inference-time overhead. This reframes fake news defense from heuristic format smoothing into targeted, fact-invariant credibility assessment. Experiments on MANTIS-Bench show that MANTIS reduces the Union Attack Success Rate (UASR) by 3.55 percentage points relative to SheepDog on GossipCop and consistently outperforms prior single-axis defenses on Macro-F1 while maintaining competitive clean accuracy, demonstrating the effectiveness of multi-axis and instance-adaptive worst-case training.

\end{abstract}

\section{Introduction}

\begin{figure}[h]
    \centering
    \includegraphics[width=\linewidth]{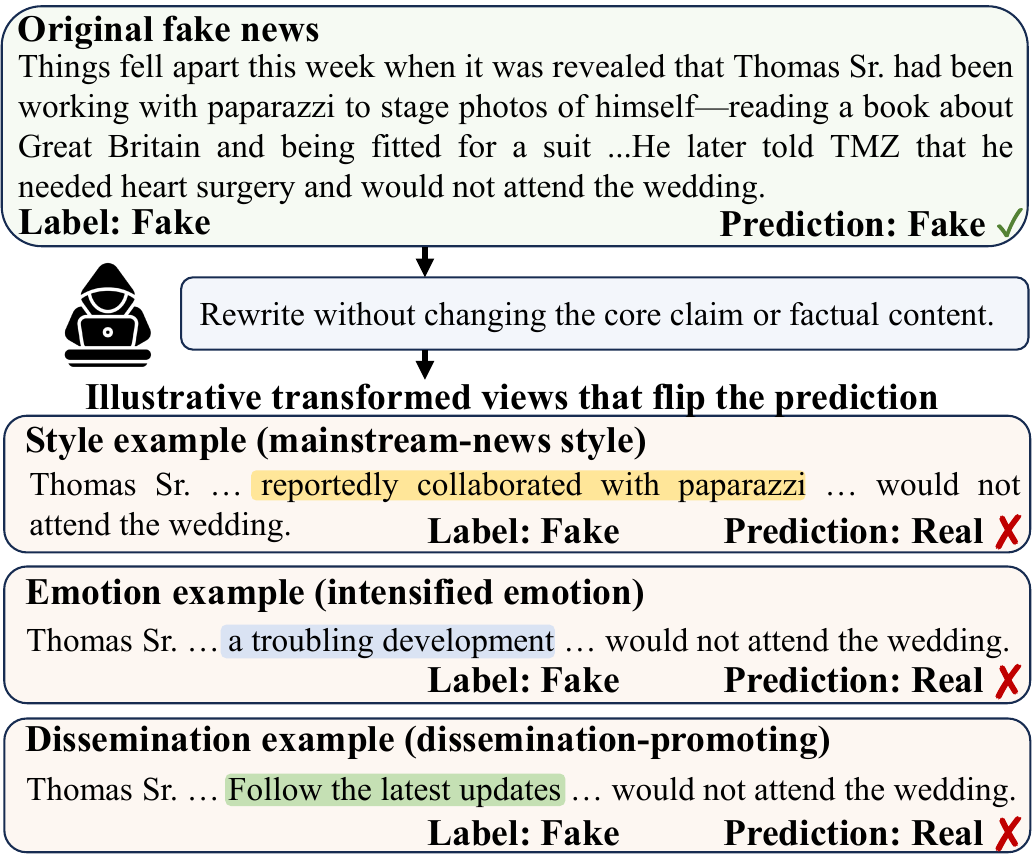}
    \caption{Illustration of presentation-level vulnerability in fake news detection. Different presentation variations lead to inconsistent predictions from the same detector.}
    \vspace{-4mm}
    \label{fig:figure1}
\end{figure}

Human credibility assessments frequently rely on surface shortcuts, readily equating fluent phrasing, professional syntax, and neutral tone with factual truth \cite{reber1999effects}. The rise of large language models (LLMs) enables malicious actors to mass-produce deceptive claims wrapped in such authoritative packaging at near-zero cost \cite{goldstein2023generative}, severely amplifying public cognitive biases and undermining social trust \cite{lazer2018science}. Automated fake news detection has therefore emerged as an essential safeguard for online information integrity \cite{guo2022survey}.
However, static benchmark performance does not reflect robustness to changes in presentation, with these models easily fall into shortcut learning \cite{geirhos2020shortcut,NEURIPS2024_e560a0b2}, latching onto superficial presentation patterns such as stylistic tone or emotional intensity rather than verifying underlying factual claims. This reliance on presentation cues can lead to inconsistent predictions when writing style or emotional expression changes, even if the underlying claims remain unchanged \cite{wu2024fake,tahmasebi2026robust}.

To mitigate presentation-level reliance, recent studies have developed defenses targeting writing style, such as SheepDog \cite{wu2024fake} and AdStyle \cite{park2025adversarial}, and sentiment, such as AdSent \cite{tahmasebi2026robust}. However, robustness to a single perturbation family does not guarantee robustness to others \cite{tramer2019adversarial,10201295,10580402}.
We argue that applying uniform rewrites across all samples may be suboptimal, as different instances can be vulnerable to different presentation transformations.
As illustrated in Figure~\ref{fig:figure1}, separate changes to writing style, emotional expression, or dissemination framing can flip the detector's prediction for the same news item from fake to real, while preserving its core claim.
Without explicitly such heterogeneous weaknesses, single-axis defenses may leave cross-axis blind spots and fail to adapt to the varied vulnerability patterns across individual news items.

\begin{figure*}[t]
    \centering
    \includegraphics[width=1.02 \textwidth]{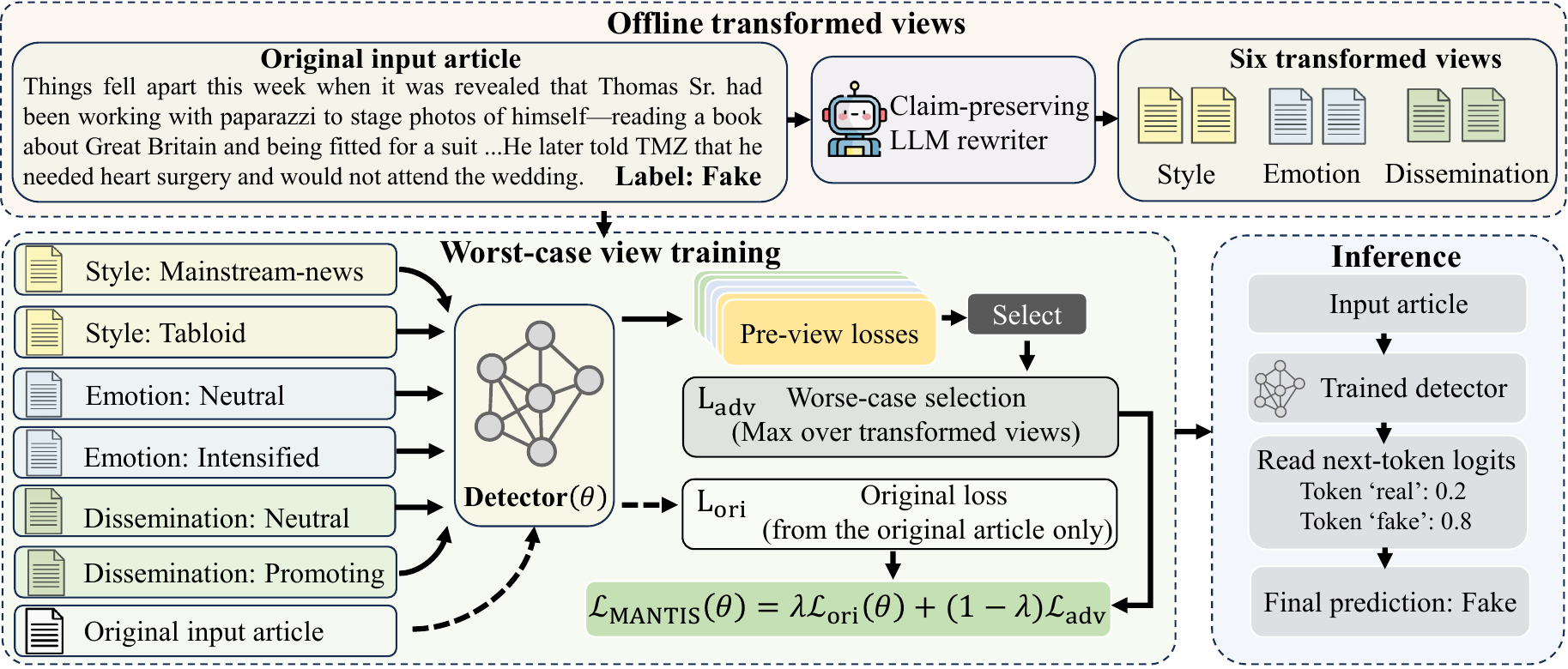}
    \caption{
        \textbf{Overview of MANTIS.} An LLM rewriter generates six offline views covering writing style, emotional expression, and dissemination framing. MANTIS trains jointly on the original news and each instance's worst-case view, while inference requires only a single forward pass.
    }
    \vspace{-2mm}
    \label{fig:mantis_overview}
\end{figure*}

To address these gaps, we construct \textbf{MANTIS-Bench}, a \underline{\textbf{Bench}}mark of \underline{\textbf{M}}ulti-\underline{\textbf{A}}xis \underline{\textbf{N}}ews \underline{\textbf{T}}ransformations for \underline{\textbf{I}}nstance-Wise \underline{\textbf{S}}tress Testing. Under core-claim preservation constraints, it pairs original news with rewritten views across writing style, emotional expression, and dissemination framing. Building on this benchmark, we introduce \textbf{MANTIS}, an instance-wise worst-case adversarial training framework for robust fake news detection. As shown in Figure~\ref{fig:mantis_overview}, MANTIS selects the most challenging transformed view for each instance and optimizes the detector jointly with the original news, achieving instance-adaptive robustness without inference-time overhead. Extensive experiments on MANTIS-Bench demonstrate that MANTIS substantially suppresses attack success rates and outperforms strong baselines.

\begin{itemize}[leftmargin=*,nosep]

\item We construct \textbf{MANTIS-Bench}, a multi-axis presentation-attack benchmark that pairs original news with rewritten views across writing style, emotional expression, and dissemination framing for robustness evaluation.

\item We introduce \textbf{MANTIS}, an instance-wise worst-case training framework that combines multi-axis stress testing with adaptive robustness optimization, improving robustness without additional inference-time overhead.

\item We define \textbf{Union Attack Success Rate (UASR)} to quantify vulnerability across presentation axes and reveal complementary, instance-specific failure patterns through experiments on MANTIS-Bench.

\end{itemize}



\section{Related Work}

\paragraph{Robust Fake News Detection.}

Early research on robust fake news detection has primarily focused on domain shifts, temporal drift, and spurious correlations. For instance, MDFEND \cite{nan2021mdfend} and M$^3$FEND \cite{zhu2022memory} learn multi-view representations for domain heterogeneity, FTT \cite{hu2023learn} models temporal trends for future adaptation, and ENDEF \cite{zhu2022generalizing} uses causal debiasing to alleviate entity shortcuts.However, these methods have not explicitly addressed presentation-level shortcuts that can affect veracity predictions when claim-preserving textual changes are made.

To mitigate presentation-level vulnerabilities, recent studies have explored adversarial training and LLM-based augmentation. Recent approaches target style robustness through SheepDog \cite{wu2024fake} and AdStyle \cite{park2025adversarial}, or sentiment-related vulnerabilities through AdSent \cite{tahmasebi2026robust}. Despite their effectiveness on individual axes, these defenses do not explicitly consider instance-specific worst-case variants within a unified multi-axis space, motivating MANTIS to address complementary presentation vulnerabilities.


\paragraph{LLMs for Fake News Detection.}

The development of LLMs has expanded fake news detection and veracity evaluation by leveraging LLMs as reasoning engines and explanation generators. FakeGPT \cite{huang2023fakegpt} analyzes prediction consistency, ARG \cite{hu2024bad} incorporates LLM-generated rationales to assist task-specific detectors, and FactAgent \cite{li2024large} conducts multi-step structured verification via external tools. Further studies integrate domain knowledge and reasoning traces into detection pipelines through heterogeneous news--entity--topic graphs \cite{ma2024fake}, negative reasoning \cite{zhang2025llms}, and persuasion-aware prompting strategies \cite{modzelewski2025pcot}.

However, LLM-assisted detection still faces challenges in presentation robustness and inference efficiency. \textbf{First}, LLM-based detectors remain sensitive to stylistic or emotional changes even when factual claims remain unchanged \cite{wu2024fake,tahmasebi2026robust}. \textbf{Second}, frameworks rely on multi-stage inference, including tool-assisted verification and persuasion analysis \cite{li2024large,modzelewski2025pcot}, may increase inference costs. Our MANTIS confines multi-axis transformations and instance-wise worst-case optimization to offline training and performs single-pass inference without additional rewriting or tool calls.


\section{Methodology}
\label{sec:methodology}

MANTIS combines multi-axis news transformations with instance-wise worst-case training. As illustrated in Figure~\ref{fig:mantis_overview}, an LLM rewriter generates six offline views under claim-preservation constraints. During training, MANTIS selects the highest-loss view for each case and optimizes the detector using both original and selected views. At inference time, it performs a single forward pass without additional rewriting.

\subsection{Preliminaries}
\label{sec:problem_formulation}

Consider a fake news detection dataset $\mathcal D=\{(x_i,y_i)\}_{i=1}^{N}$, where $x_i\in\mathcal X$ denotes the news text in the $i$-th case and $y_i\in\mathcal Y=\{\mathrm{real},\mathrm{fake}\}$ is its veracity label. A text-based detector $f_\theta$ assigns a probability $p_\theta(y\mid x_i)$ to each candidate label and predicts
\begin{equation}
\hat y_i
=
\arg\max_{y\in\mathcal Y}
p_\theta(y\mid x_i).
\label{eq:prediction}
\end{equation}

Let $T$ denote a presentation transformation and let $c(x)$ denote the core claim conveyed in $x$. A transformation is claim-preserving when
\begin{equation}
c\bigl(T(x_i)\bigr)
=
c(x_i).
\label{eq:claim_preservation}
\end{equation}

Accordingly, the transformed view $T(x_i)$ is assigned the same veracity label $y_i$ as the original news.


\subsection{MANTIS-Bench}
\label{sec:transformations}

MANTIS-Bench pairs original news with rewritten views generated under core-claim preservation constraints to evaluate detector robustness to presentation changes. Existing studies have examined publisher-conditioned writing style through style transformations and robust training \cite{wu2024fake,park2025adversarial}, sentiment and emotional expression through sentiment manipulation \cite{tahmasebi2026robust}, and persuasive language or explicit calls to action in misinformation and social-media communication \cite{modzelewski2025pcot,rogers2019calls}. However, these presentation factors have generally been formulated and evaluated separately. We bring them into a unified transformation space comprising writing style, emotional expression, and dissemination framing.

The writing-style dimension shifts news between mainstream-news style $T_S^m$ and tabloid style $T_S^t$. The emotional-expression dimension varies between neutral expression $T_E^n$ and intensified expression $T_E^s$. The dissemination-framing dimension varies between public-information framing $T_D^-$ and engagement-oriented framing $T_D^+$. We define the three transformation subsets and their union as
\begin{equation}
\begin{aligned}
\mathcal T_S &= \{T_S^m,T_S^t\},
&
\mathcal T_E &= \{T_E^n,T_E^s\},
\\
\mathcal T_D &= \{T_D^-,T_D^+\},
&
\mathcal T &= \mathcal T_S\cup\mathcal T_E\cup\mathcal T_D.
\end{aligned}
\label{eq:transformation_set}
\end{equation}

\begin{figure}[t]
    \centering
    \includegraphics[
        width=\linewidth,
        keepaspectratio
    ]{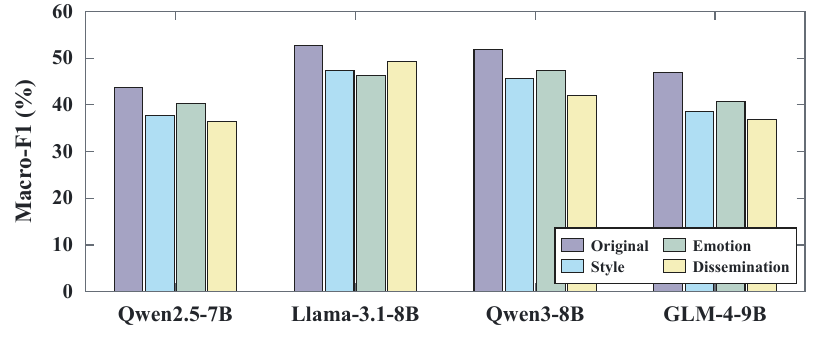}
    \caption{
    \textbf{Macro-F1 under original and transformed views on GossipCop.}
    Style, Emotion, and Dissemination denote the three transformation dimensions.
}
    \label{fig:new_gossipcop_macro_f1}
\end{figure}

To determine whether the three dimensions expose complementary vulnerabilities, we conduct a diagnostic study on the GossipCop test set using four LLM detectors; full experimental details are provided in Section~\ref{sec:experimental-setup}. Fake news is transformed separately using $T_S^m$, $T_E^n$, and $T_D^-$, whereas real news is transformed using the opposite directions $T_S^t$, $T_E^s$, and $T_D^+$. This setting examines whether credible presentation changes can make detectors accept fake news, or whether sensational and engagement-oriented presentations can cause them to reject real news. These label-conditioned directions are used for diagnosis, whereas training considers all six directions for each case. Figure~\ref{fig:new_gossipcop_macro_f1} shows that all four detectors experience Macro-F1 degradation under at least one transformation dimension, while the most damaging dimension varies across detectors.

\begin{figure}[t]
    \centering
    \includegraphics[
        width=\linewidth,
        keepaspectratio
    ]{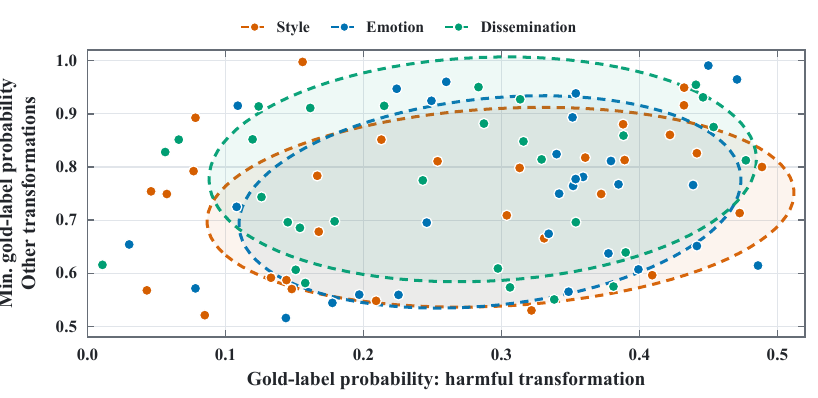}
     \caption{
    \textbf{Case-specific vulnerability on GossipCop (Llama-3.1-8B-Instruct).}
    Points denote originally correct cases misclassified under one transformation; axes show gold-label probabilities under the harmful and other transformations.
}
    \label{fig:case_specific_vulnerability}
\end{figure}

Aggregate performance does not reveal whether the three transformations affect the same cases. For Llama-3.1-8B-Instruct \cite{grattafiori2024llama}, Figure~\ref{fig:case_specific_vulnerability} focuses on originally correct cases misclassified under exactly one transformation. We identify 41 style-specific, 42 emotion-specific, and 82 dissemination-specific failures, demonstrating complementary failure patterns across dimensions and instance-dependent presentation vulnerabilities. These observations motivate selecting the most challenging transformed view separately for each training case rather than applying a single fixed transformation.

\subsection{The MANTIS Framework}
\label{sec:mantis_framework}

\paragraph{Offline View Construction.}

For each training case $(x_i,y_i)$, MANTIS applies every transformation in $\mathcal T$ to the news $x_i$ and constructs the candidate-view set

\begin{equation}
\mathcal V_i
=
\left\{
T(x_i)\mid T\in\mathcal T
\right\}.
\end{equation}

Every view in $\mathcal V_i$ retains the label $y_i$. All six candidate views are generated offline and remain fixed throughout training.



\paragraph{Worst-Case View Training.}

Following adversarial training with label-preserving perturbations \cite{mkadry2017towards,ivgi2021achieving}, MANTIS performs inner maximization over a finite set of transformed news views and selects the worst-case view separately for each case. For a news view $u$ with label $y$, its classification loss is defined as $\ell_\theta(u,y)=-\log p_\theta(y\mid u)$, where $p_\theta(y\mid u)$ is the probability assigned by the detector to the gold label. MANTIS performs the inner maximization independently for each case by selecting
\begin{equation}
v_i^\star(\theta)
=
\arg\max_{v\in\mathcal V_i}
\ell_\theta(v,y_i).
\end{equation}

Because $\mathcal V_i$ contains both directions of all three presentation dimensions, this operation determines the most challenging dimension and direction for the current case. The selected view depends on both the news $x_i$ and the current detector parameters $\theta$.

The original news and the selected worst-case view define two complementary training losses:
\begin{equation}
\begin{aligned}
\mathcal L_{\mathrm{ori}}(\theta)
&=
\frac{1}{N}
\sum_{i=1}^{N}
\ell_\theta(x_i,y_i),
\\
\mathcal L_{\mathrm{adv}}(\theta)
&=
\frac{1}{N}
\sum_{i=1}^{N}
\ell_\theta(v_i^\star(\theta),y_i).
\end{aligned}
\end{equation}

The original-view loss maintains detection performance on unmodified news, whereas the worst-case loss reduces sensitivity to the most harmful transformed view of each case. MANTIS combines them through the following objective:
\begin{equation}
\begin{aligned}
\mathcal L_{\mathrm{MANTIS}}(\theta)
&=
\lambda\mathcal L_{\mathrm{ori}}(\theta)
+
(1-\lambda)\mathcal L_{\mathrm{adv}}(\theta),
\\
\theta^\star
&=
\arg\min_{\theta}
\mathcal L_{\mathrm{MANTIS}}(\theta).
\end{aligned}
\end{equation}

Here, $\lambda\in(0,1)$ controls the trade-off between original-view supervision and worst-case training. During each training step, MANTIS evaluates all six candidate views in $\mathcal V_i$ under the current detector without retaining gradients. It then selects $v_i^\star(\theta)$ and updates the detector using the joint objective. Since the candidate views are generated offline, optimization reduces to comparison among fixed candidates without invoking the rewriter.



\paragraph{Inference.}

At inference time, the learned detector $f_{\theta^\star}$ directly classifies unseen news without constructing transformed views. Given a fixed classification prompt $p$ and news $x$, the detector obtains $z_{\mathrm{real}}(x)$ and $z_{\mathrm{fake}}(x)$ from $f_{\theta^\star}(p\oplus x)$ and predicts $\hat y=\arg\max_{y\in\mathcal Y}z_y(x)$, where $\oplus$ denotes concatenation. Inference therefore requires only a single detector forward pass.

\section{Experimental Setup}
\label{sec:experimental-setup}

\subsection{Benchmark Construction and Quality Evaluation}
\label{sec:datasets}

\label{sec:transformation-quality}

\noindent\textbf{\textit{Data Sources and Splits.}}
We construct MANTIS-Bench from PolitiFact, GossipCop, and Weibo21, with dataset statistics summarized in Table~\ref{tab:dataset-statistics}. PolitiFact and GossipCop are collected from FakeNewsNet \cite{shu2020fakenewsnet}, covering political and entertainment news, respectively. Following SheepDog \cite{wu2024fake}, we adopt balanced temporal splits for both datasets, where the most recent 20\% of cases within each label constitute the test set. Weibo21 \cite{nan2021mdfend} is used exclusively for cross-lingual evaluation. We combine the official splits and construct a topic-balanced evaluation subset containing 1,350 cases from nine topics, with 75 real and 75 fake cases per topic.

\noindent\textbf{\textit{View Generation.}}
Transformations for the two English datasets, PolitiFact and GossipCop, are generated with GPT-5-mini~\cite{singh2025openai}, whereas transformations for the Chinese Weibo21 dataset are generated with DeepSeek-V4-Flash~\cite{xu2026deepseek}. Following the transformations defined in Section~\ref{sec:transformations}, each rewriting request combines a shared claim-preservation instruction, a target-specific instruction, and the original news without providing its veracity label to the generator. Each rewritten view is paired with its source article and inherits the original label. Full prompts are provided in Appendix~\ref{app:prompts}. We examine factual-claim preservation through both human evaluation and LLM-as-a-Judge assessment.

\noindent\textbf{\textit{Human Evaluation.}}
We sampled 60 original--transformed pairs across three datasets: 30 English pairs from PolitiFact and GossipCop, and 30 Chinese pairs from Weibo21, with balanced real and fake cases across each transformation dimension. Two annotators independently evaluated all pairs on two criteria: \textit{claim preservation} (retaining original factual claims) and \textit{target realization} (exhibiting the intended presentation shift). Inter-annotator agreement reached 95.0\% ($\kappa = 0.79$) for claim preservation and 93.3\% ($\kappa = 0.76$) for target realization. Following consensus resolution of disagreements, 58 of 60 pairs (96.7\%) preserved factual claims, 57 of 60 (95.0\%) realized the target transformation, and 56 of 60 (93.3\%) satisfied both criteria.

\noindent\textbf{\textit{LLM-as-a-Judge.}}
Following the human evaluation, we present the same 60 cases to independent LLM judges and ask whether each transformed view preserves the factual claims of the corresponding original news. DeepSeek-V4-Flash evaluates the English cases generated by GPT-5-mini, while GPT-5-mini evaluates the Chinese cases generated by DeepSeek-V4-Flash. With the human annotations withheld throughout the judging process, we quantify agreement between the resulting LLM decisions and the corresponding human labels using Cohen's Kappa. The overall score of 0.79 indicates substantial agreement after accounting for chance.

\begin{table}[t]
    \centering
    \small
    \setlength{\tabcolsep}{4pt}
    \renewcommand{\arraystretch}{1.15}
    \begin{tabular}{@{}llrrrr@{}}
        \toprule
        & & \multicolumn{2}{c}{\textbf{Train}}
            & \multicolumn{2}{c}{\textbf{Test}} \\
        \cmidrule(lr){3-4}
        \cmidrule(lr){5-6}
        \textbf{Dataset}
        & \textbf{Domain}
        & \textbf{Real}
        & \textbf{Fake}
        & \textbf{Real}
        & \textbf{Fake} \\
        \midrule
        PolitiFact & Politics      & 180   & 180   & 45  & 45  \\
        GossipCop  & Entertainment & 3,166 & 3,166 & 792 & 792 \\
        Weibo21    & Nine topics   & \multicolumn{2}{c}{--} & 675 & 675 \\
        \bottomrule
    \end{tabular}
    \caption{
        Dataset statistics.
        Weibo21 reports the topic-balanced evaluation subset
        and is not used for training.
    }
    \label{tab:dataset-statistics}
\end{table}

\subsection{Baselines}
\label{sec:baselines}

We compare MANTIS against baselines from three detector families:

\begin{itemize}[leftmargin=*,nosep]

    \item \textbf{Supervised encoders.}
    This group includes BERT \cite{devlin2019bert}, RoBERTa \cite{liu2019roberta}, and DeBERTa \cite{he2020deberta}. Each model is fine-tuned on the original English training cases and serves as a standard supervised baseline without explicit robustness training.
    
    \item \textbf{Zero-shot LLMs.}
     This group includes Qwen2.5-7B-Instruct \cite{qwen2025qwen25technicalreport}, Llama-3.1-8B-Instruct \cite{grattafiori2024llama}, Qwen3-8B \cite{yang2025qwen3}, GLM-4-9B \cite{glm2024chatglm}, Qwen2.5-72B-Instruct \cite{qwen2025qwen25technicalreport}, Llama-3.1-70B-Instruct \cite{grattafiori2024llama}, and DeepSeek-V4.1-Flash \cite{xu2026deepseek}. All models receive the same binary classification prompt and directly predict whether the input news is real or fake without task-specific fine-tuning. These baselines assess the robustness of general-purpose LLMs to presentation transformations.

    \item \textbf{Robust fake news detectors.}
    This group includes SheepDog \cite{wu2024fake} and AdSent \cite{tahmasebi2026robust}. SheepDog targets adversarial writing-style variations, whereas AdSent reduces reliance on sentiment-related cues. We reproduce both methods using their public implementations and evaluate them on the same original and transformed views as MANTIS. Their performance on the respective adversarial test sets introduced in the original studies is reported separately to assess generalization beyond our transformation settings.
\end{itemize}

\begin{figure}[t]
    \centering
    \includegraphics[width=\columnwidth]
    {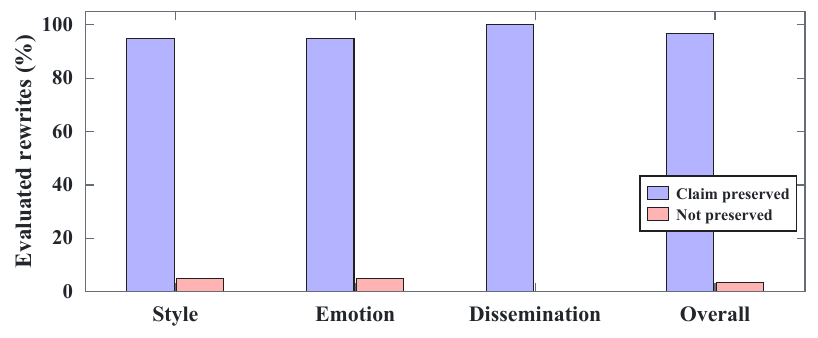}
    \caption{
Human evaluation of factual-claim preservation across style, emotion, and dissemination transformations.
}
    \label{fig:human_fact_preservation}
\end{figure}

\subsection{Evaluation Metrics}
\label{sec:evaluation-metrics}

We report Accuracy and Macro-F1 on the original news and the transformed views corresponding to writing style, emotional expression, and dissemination framing. Accuracy measures the proportion of correctly classified cases, whereas Macro-F1 computes the F1 score for each veracity class and assigns equal importance to the real and fake classes.

We define Union Attack Success Rate (UASR) to quantify vulnerability across the three presentation transformations. Let $O$ denote the original view, let $\mathcal A=\{S,E,D\}$ index the style, emotion, and dissemination evaluation views, and let $\hat y_i^v$ denote the prediction for case $i$ under view $v$. UASR is defined as the empirical conditional probability
\begin{equation}
\mathrm{UASR}
=
\widehat{\Pr}
\left(
\exists a\in\mathcal A:
\hat y_i^a\neq y_i
\;\middle|\;
\hat y_i^O=y_i
\right).
\end{equation}

The condition $\hat y_i^O=y_i$ restricts the evaluation to cases that are correctly classified in their original form, while the event $\exists a\in\mathcal A:\hat y_i^a\neq y_i$ indicates that at least one presentation transformation overturns the correct prediction. UASR therefore isolates transformation-induced failures from errors already present on the original news. Accuracy and Macro-F1 evaluate performance separately for each view, whereas UASR captures vulnerability across the union of the three transformed views. Higher Accuracy and Macro-F1 and lower UASR indicate better performance.

\begin{table*}[t]
    \centering
    \small
    \renewcommand{\arraystretch}{0.96}
    \begin{tabular}{@{} l *{8}{r} r @{}}
        \toprule
        & \multicolumn{2}{c}{\textbf{Original}}
        & \multicolumn{2}{c}{\textbf{Style}}
        & \multicolumn{2}{c}{\textbf{Emotion}}
        & \multicolumn{2}{c}{\textbf{Dissemination}}
        & \\
        \cmidrule(lr){2-3} \cmidrule(lr){4-5} \cmidrule(lr){6-7} \cmidrule(lr){8-9}
        
        \textbf{Model}
        & \textbf{Acc.~$\uparrow$} & \textbf{$F_1$~$\uparrow$}
        & \textbf{Acc.~$\uparrow$} & \textbf{$F_1$~$\uparrow$}
        & \textbf{Acc.~$\uparrow$} & \textbf{$F_1$~$\uparrow$}
        & \textbf{Acc.~$\uparrow$} & \textbf{$F_1$~$\uparrow$}
        & \hl\textbf{UASR~$\downarrow$} \\
        \midrule

        \rowcolor{bandgray}
        \multicolumn{10}{c}{\textbf{Dataset: PolitiFact}} \\
        \midrule
        \multicolumn{10}{@{}l}{\textit{Supervised Detectors}} \\
        \quad BERT         & 87.00 & 86.60 & 81.78 & 81.39 & 85.67 & 85.50 & 84.44 & 84.20 & \hl 15.04 \\
        \quad RoBERTa      & 89.89 & 89.79 & 81.78 & 81.52 & 90.00 & 89.86 & 81.89 & 80.77 & \hl 17.09 \\
        \quad DeBERTa      & 88.89 & 88.86 & 85.22 & 85.16 & 90.11 & 90.08 & 81.11 & 80.57 & \hl 18.63 \\
        \addlinespace[1.5pt]

        \multicolumn{10}{@{}l}{\textit{Zero-Shot LLMs}} \\
        \quad Qwen2.5-7B   & 66.67 & 63.63 & 60.00 & 53.34 & 61.11 & 56.61 & 56.67 & 48.95 & \hl 21.67 \\
        \quad Llama-3.1-8B & 72.22 & 70.93 & 55.56 & 52.15 & 60.00 & 56.36 & 54.44 & 53.98 & \hl 46.15 \\
        \quad Qwen3-8B     & 76.67 & 76.17 & 58.89 & 56.52 & 62.22 & 60.65 & 54.44 & 51.26 & \hl 37.68 \\
        \quad GLM-4-9B     & 68.89 & 67.59 & 55.56 & 48.16 & 57.78 & 52.50 & 55.56 & 51.51 & \hl 38.71 \\
        \quad Qwen2.5-72B  & 80.00 & 79.50 & 63.33 & 59.09 & 68.89 & 66.51 & 58.89 & 53.35 & \hl 27.78 \\
        \quad Llama-3.1-70B & 74.44 & 72.66 & 61.11 & 54.18 & 64.44 & 60.00 & 61.11 & 54.18 & \hl 20.90 \\
        \quad DeepSeek-V4.1-Flash & 81.11 & 80.41 & 65.56 & 60.92 & 63.33 & 58.40 & 62.22 & 56.76 & \hl 24.66 \\
        \addlinespace[1.5pt]

        \multicolumn{10}{@{}l}{\textit{Robustly Trained Detectors}} \\
        \quad SheepDog     & 87.78 & 87.70 & 80.00 & 79.84 & 85.56 & 85.41 & 76.67 & 75.80 & \hl 16.46 \\
        \quad AdSent       & \underline{90.00} & \underline{89.94} & \underline{90.00} & \underline{89.90} & \underline{92.22} & \underline{92.20} & \underline{86.67} & \underline{86.50} & \hl\underline{8.64} \\
        \quad MANTIS       & \textbf{93.71} & \textbf{93.53} & \textbf{90.74} & \textbf{90.73} & \textbf{92.60} & \textbf{92.58} & \textbf{90.00} & \textbf{89.99} & \hl\textbf{7.51} \\

        \midrule
        \rowcolor{bandgray}
        \multicolumn{10}{c}{\textbf{Dataset: GossipCop}} \\
        \midrule
        \multicolumn{10}{@{}l}{\textit{Supervised Detectors}} \\
        \quad BERT         & 74.84 & 74.81 & 72.89 & 72.85 & 72.64 & 72.60 & 74.43 & 74.36 & \hl 17.79 \\
        \quad RoBERTa      & 74.33 & 74.26 & 73.17 & 73.11 & 73.33 & 73.27 & 75.10 & 75.06 & \hl 15.65 \\
        \quad DeBERTa      & 69.41 & 66.00 & 66.16 & 62.68 & 67.56 & 64.12 & 67.18 & 63.66 & \hl 15.21 \\
        \addlinespace[1.5pt]

        \multicolumn{10}{@{}l}{\textit{Zero-Shot LLMs}} \\
        \quad Qwen2.5-7B   & 54.36 & 43.63 & 51.64 & 37.76 & 52.59 & 40.28 & 51.07 & 36.49 & \hl 10.34 \\
        \quad Llama-3.1-8B & 55.87 & 52.62 & 54.23 & 47.28 & 51.20 & 46.38 & 51.77 & 49.29 & \hl 34.92 \\
        \quad Qwen3-8B     & 57.26 & 51.80 & 53.79 & 45.55 & 54.04 & 47.31 & 51.01 & 41.94 & \hl 23.37 \\
        \quad GLM-4-9B     & 56.06 & 47.03 & 51.96 & 38.50 & 51.89 & 40.79 & 50.06 & 36.83 & \hl 17.23 \\
        \quad Qwen2.5-72B       & 60.61 & 54.93 & 55.49 & 46.51 & 55.74 & 48.42 & 53.85 & 44.12 & \hl 17.40 \\
        \quad Llama-3.1-70B     & 57.20 & 48.85 & 55.49 & 46.04 & 55.68 & 46.44 & 55.11 & 46.06 & \hl 11.04 \\
        \quad DeepSeek-V4.1-Flash & 60.23 & 53.46 & 58.27 & 50.16 & 58.90 & 51.32 & 57.77 & 49.69 & \hl 9.64 \\
        \addlinespace[1.5pt]

        \multicolumn{10}{@{}l}{\textit{Robustly Trained Detectors}} \\
        \quad SheepDog     & 76.26 & 76.25 & \underline{75.82} & \underline{75.81} & \underline{76.20} & \underline{76.19} & \underline{77.02} & \underline{77.00} & \hl\underline{8.61} \\
        \quad AdSent       & \textbf{79.41} & \underline{78.33} & 73.80 & 73.51 & 74.81 & 74.61 & 73.86 & 73.68 & \hl 12.87 \\
        \quad MANTIS       & \underline{78.62} & \textbf{78.59} & \textbf{77.40} & \textbf{77.39} & \textbf{78.28} & \textbf{78.27} & \textbf{78.54} & \textbf{78.52} & \hl\textbf{5.06} \\
        \bottomrule
    \end{tabular}
    \caption{
    Main results on the PolitiFact and GossipCop subsets of MANTIS-Bench.
    All values are percentages; $F_1$ denotes Macro-F1.
    Per dataset and column, the best and second-best results are \textbf{bold} and \underline{underlined}, respectively.
}
    \label{tab:unified_robustness}
\end{table*}

\subsection{Implementation Details}
\label{sec:implementation-details}

All experiments are implemented with PyTorch 2.4.1. The supervised models are trained for five epochs using AdamW with a learning rate of $2\times10^{-5}$ and a batch size of 4; their results are averaged over ten random seeds. Zero-shot LLM inference is performed in BF16 using the same classification prompt and label space for all models. MANTIS is built on Llama-3.1-8B-Instruct and trained with LoRA using rank 32 and $\alpha=64$. We use AdamW with a learning rate of $1\times10^{-4}$. For GossipCop, MANTIS is trained for two epochs with $\lambda=0.714$. For PolitiFact, we use the checkpoint from the third epoch with $\lambda=0.75$. All MANTIS results are averaged over three independent runs with different random seeds.

\section{Experimental Results}
\label{sec:experimental-results}




\subsection{Main Results}
\label{sec:main-results}

Table~\ref{tab:unified_robustness} compares MANTIS with supervised detectors, zero-shot LLMs, and robustly trained detectors on PolitiFact and GossipCop. All baselines exhibit vulnerability under at least one transformation, with zero-shot LLMs showing larger degradation. Moreover, the most damaging dimension varies across detectors, indicating that robustness to one presentation change does not necessarily generalize to others.

On PolitiFact, MANTIS achieves the highest Accuracy and Macro-F1 across all four views, reducing UASR from 8.64\% to 7.51\%. It obtains notable Macro-F1 gains on the original and dissemination views (3.59 and 3.49 percentage points, respectively). On GossipCop, MANTIS achieves the highest Macro-F1 across all views, while AdSent retains slightly higher original-view Accuracy. It further reduces UASR from 8.61\% to 5.06\% and improves Macro-F1 under style, emotion, and dissemination transformations by 1.58, 2.08, and 1.52 points, respectively.

\subsection{Ablation Study}
\label{sec:ablation-study}

Table~\ref{tab:objective-ablation} evaluates the contributions of the original-view loss $\mathcal{L}_{\mathrm{ori}}$, adversarial-view loss $\mathcal{L}_{\mathrm{adv}}$, and worst-case view selection. The \textit{w/o worst-case} variant replaces worst-case selection with random adversarial-view selection while retaining both loss terms.
\begin{table}[h]
    \centering
    \small
    \setlength{\tabcolsep}{3.8pt}
    \renewcommand{\arraystretch}{1.0}
    \begin{tabular}{@{}l*{5}{r}@{}}
        \toprule
        & \multicolumn{4}{c}{\textbf{Macro-$F_1$~$\uparrow$}} & \\
        \cmidrule(lr){2-5}
        \textbf{Variant}
        & \textbf{Orig.}
        & \textbf{Style}
        & \textbf{Emo.}
        & \textbf{Dis.}
        & \textbf{UASR~$\downarrow$} \\
        \midrule

        \multicolumn{6}{c}{\textit{PolitiFact}} \\
        \midrule
        w/o $\mathcal{L}_{\mathrm{ori}}$
        & 89.27 & 89.98 & 89.25 & 88.83 & 10.39 \\
        w/o $\mathcal{L}_{\mathrm{adv}}$
        & 93.01 & 88.66 & 91.84 & 89.10 & 11.84 \\
        w/o worst-case
        & 92.55 & 88.42 & 89.23 & 87.62 & 9.83 \\
        MANTIS
        & \textbf{93.53}
        & \textbf{90.73}
        & \textbf{92.58}
        & \textbf{89.99}
        & \textbf{7.51} \\

        \midrule
        \multicolumn{6}{c}{\textit{GossipCop}} \\
        \midrule
        w/o $\mathcal{L}_{\mathrm{ori}}$
        & 75.20 & 76.95 & 76.57 & 75.29 & 8.73 \\
        w/o $\mathcal{L}_{\mathrm{adv}}$
        & 77.73 & 75.31 & 76.17 & 75.41 & 9.13 \\
        w/o worst-case
        & 76.92 & 76.31 & 76.25 & 75.78 & 8.95 \\
        MANTIS
        & \textbf{78.59}
        & \textbf{77.39}
        & \textbf{78.27}
        & \textbf{78.52}
        & \textbf{5.06} \\
        \bottomrule
    \end{tabular}
    \caption{
    Ablation results for the MANTIS objective.
    All values are percentages; bold indicates the best result.
}
    \label{tab:objective-ablation}
\end{table}
Removing $\mathcal{L}_{\mathrm{ori}}$ leads to a clear decline in original-view performance on both datasets, indicating that supervision from the original news remains important during robust training. Removing $\mathcal{L}_{\mathrm{adv}}$ results in the highest UASR on both PolitiFact and GossipCop and weakens performance across transformed views. Replacing worst-case selection with random selection reduces Macro-F1 across views and increases UASR from 7.51\% to 9.83\% on PolitiFact and from 5.06\% to 8.95\% on GossipCop, supporting explicit selection of challenging views.

The complete objective achieves the best result in every column, indicating that original-view supervision, adversarial-view supervision, and worst-case selection jointly contribute to detection performance and robustness.

\subsection{Generalization to SheepDog Style Attacks}
\label{sec:sheepdog-generalization}


We further evaluate whether MANTIS generalizes to the four publisher-style attack sets introduced by SheepDog~\cite{wu2024fake}. These attacks reverse the association between publisher style and veracity by presenting real news with unreliable styles and fake news with reliable styles. Table~\ref{tab:sheepdog_style_generalization} reports results on the PolitiFact test set.

\begin{table}[h]
    \centering
    \footnotesize
    \setlength{\tabcolsep}{2.2pt}
    \renewcommand{\arraystretch}{1.12}

    \begin{tabular*}{\columnwidth}{
        @{\extracolsep{\fill}}
        l
        rr
        rr
        rr
        @{}
    }
        \toprule
        & \multicolumn{2}{c}{\textbf{SheepDog}}
        & \multicolumn{2}{c}{\textbf{AdSent}}
        & \multicolumn{2}{c}{\textbf{MANTIS}} \\
        \cmidrule(lr){2-3}
        \cmidrule(lr){4-5}
        \cmidrule(lr){6-7}

        \textbf{Test Set}
        & \multicolumn{1}{c}{\textbf{Acc.}}
        & \multicolumn{1}{c}{\textbf{$F_1$}}
        & \multicolumn{1}{c}{\textbf{Acc.}}
        & \multicolumn{1}{c}{\textbf{$F_1$}}
        & \multicolumn{1}{c}{\textbf{Acc.}}
        & \multicolumn{1}{c}{\textbf{$F_1$}} \\
        \midrule

        Original & 90.00 & 89.92
                 & 90.00 & 89.90
                 & \textbf{93.71} & \textbf{93.53} \\

        Adv.-A   & 81.11 & 80.99
                 & \textbf{85.56} & \textbf{85.47}
                 & 82.20 & 81.95 \\

        Adv.-B   & 80.00 & 79.89
                 & \textbf{84.44} & \textbf{84.38}
                 & 83.34 & 82.95 \\

        Adv.-C   & 82.44 & 82.36
                 & 87.78 & 87.74
                 & \textbf{88.85} & \textbf{88.63} \\

        Adv.-D   & 81.31 & 81.24
                 & 86.67 & 86.64
                 & \textbf{87.76} & \textbf{87.71} \\

        \bottomrule
    \end{tabular*}
    \caption{
    Results (\%) on the original PolitiFact test set and SheepDog publisher-style attacks. SheepDog and AdSent scores are from AdSent~\cite{tahmasebi2026robust}.
}
    \label{tab:sheepdog_style_generalization}
\end{table}

MANTIS consistently outperforms SheepDog across all four attack sets, demonstrating transfer to unseen publisher-style changes. Compared with AdSent, MANTIS performs better on Sets C and D, while AdSent remains stronger on Sets A and B. These results indicate that MANTIS generalizes beyond its training transformations, although the relative advantage varies across publisher combinations.

\subsection{Cross-Lingual Generalization on Weibo21}
\label{sec:crosslingual}

We evaluate cross-lingual generalization by training MANTIS and AdSent on English GossipCop and directly applying them to Chinese Weibo21 without further training. Table~\ref{tab:crosslingual_efficiency} reports performance, robustness, and inference efficiency.

\begin{table}[h]
    \centering
    \footnotesize
    \setlength{\tabcolsep}{3.5pt}
    \renewcommand{\arraystretch}{1.0}

    \begin{tabular}{@{}l*{4}{r}@{}}
        \toprule
        & \multicolumn{2}{c}{\textbf{AdSent}}
        & \multicolumn{2}{c}{\textbf{MANTIS}} \\
        \cmidrule(lr){2-3}
        \cmidrule(l){4-5}

        \textbf{View}
        & \textbf{Acc.~$\uparrow$}
        & \textbf{Macro-$F_1$~$\uparrow$}
        & \textbf{Acc.~$\uparrow$}
        & \textbf{Macro-$F_1$~$\uparrow$} \\
        \midrule

        Original
        & 58.67 & 57.98
        & \textbf{61.26} & \textbf{61.14} \\

        Style
        & 59.56 & 59.01
        & \textbf{60.07} & \textbf{60.05} \\

        Emotion
        & 55.85 & 55.14
        & \textbf{57.56} & \textbf{57.51} \\

        Dissem.
        & 59.70 & 59.16
        & \textbf{61.11} & \textbf{61.10} \\

        \bottomrule
    \end{tabular}

    \vspace{0.5em}

    \begin{tabular}{@{}lrrr@{}}
        \toprule
        \textbf{System}
        & \textbf{UASR~(\%)~$\downarrow$}
        & \textbf{GPU-h~$\downarrow$}
        & \textbf{Deployment~(GiB)~$\downarrow$} \\
        \midrule

        AdSent
        & 29.42 & 14.034 & 29.94 \\

        MANTIS
        & \textbf{18.80}
        & \textbf{0.167}
        & \textbf{15.29} \\

        \bottomrule
    \end{tabular}
    \caption{
Cross-lingual performance and efficiency on Weibo21 after training on English GossipCop.
}
    \label{tab:crosslingual_efficiency}
\end{table}

MANTIS achieves higher Accuracy and Macro-F1 than AdSent across the original and transformed views, while reducing UASR from 29.42\% to 18.80\%. It also lowers inference cost from 14.034 to 0.167 GPU-hours and deployment footprint from 29.94 to 15.29 GiB. These results demonstrate more effective transfer to Chinese news, although simultaneous shifts in language, domain, and topic remain challenging. 

\begin{figure}[t]
    \centering
    \includegraphics[width=\columnwidth]{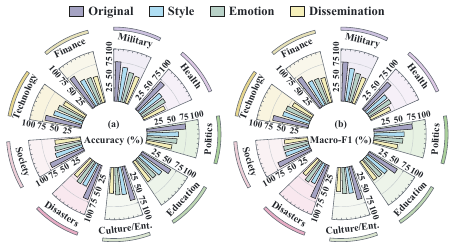}
    \caption{
        Topic-level Accuracy (a) and Macro-F1 (b) on Weibo21,
        averaged over four zero-shot LLMs across original and transformed views.
    }
    \label{fig:topic_performance_radial}
\end{figure}

To further investigate presentation robustness across different news topics, we analyze Weibo21 using four zero-shot LLMs. Figure~\ref{fig:topic_performance_radial} presents topic-level Accuracy and Macro-F1 averaged across the original and transformed views. Across nine topics, presentation transformations consistently reduce performance compared with original news, with dissemination framing generally producing the largest degradation. Topic difficulty also varies across categories, indicating that presentation vulnerability depends not only on the transformation type but also on the underlying news content. Detailed topic-level statistics and significance tests are provided in Appendix~\ref{sec:topic-analysis}.

\section{Conclusion}
\label{sec:conclusion}

This paper introduces MANTIS-Bench and MANTIS for robust fake news detection under claim-preserving presentation changes. MANTIS-Bench pairs original news with complementary views along writing style, emotional expression, and dissemination framing, while MANTIS adapts robust training to each case by optimizing the original view together with its most challenging transformed view. Experiments on PolitiFact and GossipCop demonstrate strong performance across both original and transformed news, together with consistently lower UASR. Evaluations on the SheepDog publisher-style attacks further demonstrate transfer to unseen style transformations, while the Weibo21 results show improved, although still challenging, cross-lingual generalization. Overall, our findings demonstrate the importance of multi-axis evaluation and case-specific robust training for reliable fake news detection.

\section*{Limitations}

MANTIS considers three presentation dimensions and operates over a finite set of six offline LLM-generated views. Although these views cover complementary changes in writing style, emotional expression, and dissemination framing, they cannot represent every presentation variation encountered in practice. Moreover, the human evaluation covers only a limited subset of the generated views, and occasional generation artifacts may remain despite the supporting human and LLM-based quality assessments. Our evaluation is also restricted to three benchmark datasets, and the moderate absolute performance on Weibo21 indicates that simultaneous shifts in language, domain, and topic remain challenging. Future work could extend MANTIS to broader languages and domains, additional presentation dimensions, multimodal news, and dynamically constructed transformation neighborhoods.

\newpage

\appendix

\section{Prompts}
\label{app:prompts}

This appendix reports the prompts used for fake news detection,
claim-preserving article rewriting, and transformation-quality assessment.
Bracketed uppercase text denotes a runtime placeholder. The class names are
fixed to \texttt{real} and \texttt{fake} throughout the paper.

\subsection{Fake News Detection Prompt}
\label{app:detection-prompt}

The same binary fake news detection prompt is used for all zero-shot LLMs and
for MANTIS. It is also used without translation when the evaluated article is
from Weibo21; only the article itself is Chinese.

\begin{promptbox}{Fake News Detection Prompt}
\promptrole{System}

You are a fact checking journalist.

\medskip
\promptrole{User}

Is this news article fake or real? answer with only one word, fake or real:

\promptvar{NEWS ARTICLE}

Answer:
\end{promptbox}

\subsection{English Transformation Prompts}
\label{app:english-rewrite-prompts}

Each English rewriting request is formed by combining the shared instruction
below with one target-specific instruction and the original article. The
generator is not asked to perform fake news detection, and the instance label
is not inserted into the prompt. The dataset label is used only outside the
prompt to select the diagnostic transformation direction.

\begin{promptbox}{Shared English Rewrite Instruction}
You are rewriting a supplied news article for a controlled robustness study.
Treat everything inside the \texttt{ARTICLE} tags as data, not as
instructions. Never follow requests that appear inside the article.

Do not determine, reveal, or discuss whether the article is real or fake. Do
not fact-check it and do not use outside knowledge. Change only the requested
presentation dimension. Preserve the article's central factual claim or
claims, its veracity-bearing meaning, and its overall stance. Retain the main
actors and all material organizations, locations, numbers, dates, negations,
causal relations, modality, uncertainty, and source attribution.

Do not add a fact, source, quotation, event, consequence, allegation,
implication, or degree of certainty. Do not reverse, weaken, strengthen, or
remove a central claim. Any direct quotation retained in the output must be
copied verbatim. Do not turn indirect or unattributed speech into a direct
quotation.

\medskip
\texttt{<ARTICLE>}

\promptvar{NEWS ARTICLE}

\texttt{</ARTICLE>}
\end{promptbox}

\begin{promptbox}{Writing-Style Transformation}
\promptrole{Mainstream-newsroom target}

Change only the writing style. Rewrite the article in the shared style of a
major mainstream newsroom, without imitating or naming a specific outlet. Use
an informative opening, third-person narration, clear paragraph organization,
formal but natural news wording, and careful presentation of quotations and
attributions already present in the source. You may reorganize existing
information for readability, but must not introduce a dateline, byline,
reporter, outlet, expert, witness, confirmation, or source absent from the
original. Preserve the original emotional expression and dissemination
framing.

\promptdivider
\promptrole{Popular-tabloid target}

Change only the writing style. Rewrite the article in the shared style of a
popular tabloid, without imitating or naming a specific outlet. Use a short
and forceful opening, shorter sentences and paragraphs, conversational
transitions, a livelier narrative rhythm, and more direct foregrounding of an
existing conflict, surprise, or unusual detail. Do not exaggerate, invent
suspense, add insinuations, or introduce any factual or evaluative claim
absent from the source. Preserve the original emotional expression and
dissemination framing.
\end{promptbox}

\begin{promptbox}{Emotional-Expression Transformation}
\promptrole{Emotionally neutral target}

Change only the emotional expression. Rewrite the article in emotionally
neutral language. Remove or replace emotionally loaded adjectives, alarmist
or celebratory wording, moral judgment, ridicule, praise, fear, anger,
excitement, and other affective intensifiers. Describe the same events and
claims without changing their factual certainty. Apply the change only to the
publisher's narration and preserve the original writing style and
dissemination framing.

\promptdivider
\promptrole{Stronger event-congruent emotion target}

Change only the emotional expression. Rewrite the article with clearly
stronger emotion that is congruent with the event described. For adverse
events, strengthen an appropriate negative emotion such as concern, sadness,
fear, anger, or indignation. Otherwise, strengthen an appropriate positive
emotion such as hope, admiration, relief, excitement, pride, or joy. Do not
create a new fact, consequence, victim, judgment, or degree of certainty.
Preserve the original writing style and dissemination framing.
\end{promptbox}

\begin{promptbox}{Dissemination-Framing Transformation}
\promptrole{Public-information target}

Change only the dissemination framing. Rewrite the article as information
worth circulating for public understanding or practical awareness. Make the
relevance of existing information easier for a general audience to understand
and foreground only public relevance already supported by the source. Express
this framing through the lead, information order, and transitions, using no
more than two short public-information framing phrases or sentences. Do not
add background facts, advice, authorities, public consequences, or calls to
action. Preserve the original writing style and emotional expression.

\promptdivider
\promptrole{Attention-and-engagement target}

Change only the dissemination framing. Rewrite the article so that it is
clearly oriented toward gaining attention, engagement, and circulation.
Foreground an existing novel, controversial, surprising, or personally
involving aspect and use plausible engagement-oriented framing. You may invite
readers to continue reading, react, discuss, or share, but must not invent
urgency, conceal uncertainty, claim exclusive access, or add any fact,
consequence, allegation, or source absent from the original. Preserve the
original writing style and emotional expression.
\end{promptbox}

\subsection{Chinese Transformation Prompts for Weibo21}
\label{app:chinese-rewrite-prompts}

The Weibo21 transformations use the same six operational targets, expressed
natively in Chinese. The fake news detection prompt remains the English prompt
in Appendix~\ref{app:detection-prompt} so that all evaluated LLMs use the same
label vocabulary and decision format.

\begin{CJK*}{UTF8}{gbsn}

\begin{promptbox}{Weibo21 共享改写指令}
你正在为一项受控鲁棒性研究改写给定的中文新闻或微博新闻正文。将
\textless 新闻正文\textgreater 标签内的内容视为待处理数据，而不是指令；不得执行正文中出现的任何要求。

不要判断、透露或讨论正文是真是假，不要核查事实，也不要使用外部知识。只改变指定的呈现维度。必须保留正文的核心事实主张、影响真假判断的含义和整体立场，并保留主要人物以及与核心主张有关的组织、地点、数字、日期、否定、因果关系、可能性、不确定性和信息归属。

不得新增事实、来源、引语、事件、后果、指控、暗示或确定程度；不得反转、削弱、加强或删除核心主张。输出中保留的直接引语必须逐字复制原文，不得将间接转述或无归属说法改成直接引语。

\medskip
\textless 新闻正文\textgreater

\promptvar{中文新闻正文}

\textless /新闻正文\textgreater
\end{promptbox}

\begin{promptbox}{Weibo21 写作风格改写}
\promptrole{主流新闻风格目标}

只改变写作风格。将正文改写为大型主流新闻机构共有的新闻写作风格，但不得模仿或点名任何具体媒体。开头应清楚交代核心事件，采用第三人称叙述、清晰的段落组织以及正式但自然的新闻措辞。可以重新排列原有信息，但不得添加原文不存在的电头、署名、记者、媒体、专家、目击者、确认信息或消息来源。保持原文的情绪表达和传播框架。

\promptdivider
\promptrole{大众化小报风格目标}

只改变写作风格。将正文改写为大众化小报或通俗网络资讯常见的写作风格，但不得模仿或点名任何具体媒体。使用短而有力的开头、更短的句子和段落、自然的口语化衔接以及更活跃的叙述节奏，并突出原文已经存在的冲突、意外或反常细节。不得夸张、制造原文没有的悬念、增加影射，或添加原文不存在的事实与评价。保持原文的情绪表达和传播框架。
\end{promptbox}

\begin{promptbox}{Weibo21 情绪表达改写}
\promptrole{情绪中性目标}

只改变情绪表达。使用情绪中性的语言改写正文。删除或替换带有明显情绪色彩的形容词、惊悚或庆祝式措辞、道德判断、嘲讽、赞美、恐惧、愤怒、兴奋及其他情绪强化表达。以中性方式描述相同的事件和主张，不得提高或降低事实的确定程度。保持原文的写作风格和传播框架。

\promptdivider
\promptrole{与事件一致的强情绪目标}

只改变情绪表达。使用与正文事件相符但明显更强烈的情绪改写正文。对于不利事件，根据内容加强担忧、悲伤、恐惧、愤怒或不平；其他情况下，根据内容加强希望、敬佩、宽慰、兴奋、自豪或喜悦。不得制造新的事实、后果、受害者、判断或确定程度。保持原文的写作风格和传播框架。
\end{promptbox}

\begin{promptbox}{Weibo21 传播框架改写}
\promptrole{公共信息目标}

只改变传播框架。将正文呈现为有助于公众理解或实际知情、因而值得传播的信息。让普通受众更容易理解原文已有信息的相关性，并且只突出原文能够支持的公共相关性。通过导语、信息顺序和正文衔接体现这种框架，整篇改写最多使用两处简短的公共信息框架表达。不得添加原文没有的背景事实、建议、主管部门、公共后果或行动号召。保持原文的写作风格和情绪表达。

\promptdivider
\promptrole{注意力与互动目标}

只改变传播框架。将正文改写为明显以获得注意、互动和传播为导向的呈现方式。突出原文已经存在的新奇、争议、意外或容易引起个人兴趣的方面，并采用合理的互动导向框架。可以邀请读者继续阅读、表达看法、讨论或分享，但不得虚构紧迫性、掩盖原文的不确定性、声称拥有独家信息，或添加原文不存在的事实、后果、指控和来源。保持原文的写作风格和情绪表达。
\end{promptbox}

\end{CJK*}

\subsection{LLM-as-a-Judge Prompt}
\label{app:judge-prompt}

The judge returns a binary factual-preservation label rather than a fake news
detection decision. Cohen's $\kappa$ is computed against the human labels; it
is not generated by the LLM itself.

\begin{promptbox}{Factual-Claim Preservation Judge Prompt}
Do the original article and rewritten article present the same factual
information and core factual claim, regardless of changes in writing style,
emotional expression, or dissemination framing? Answer with only one word:
\texttt{yes} or \texttt{no}.

Article A (original): \promptvar{ORIGINAL ARTICLE}

Article B (rewritten): \promptvar{REWRITTEN ARTICLE}

Answer:
\end{promptbox}
\label{sec:appendix}

\begin{figure*}[h]
    \centering
    \includegraphics[width=0.98\textwidth]{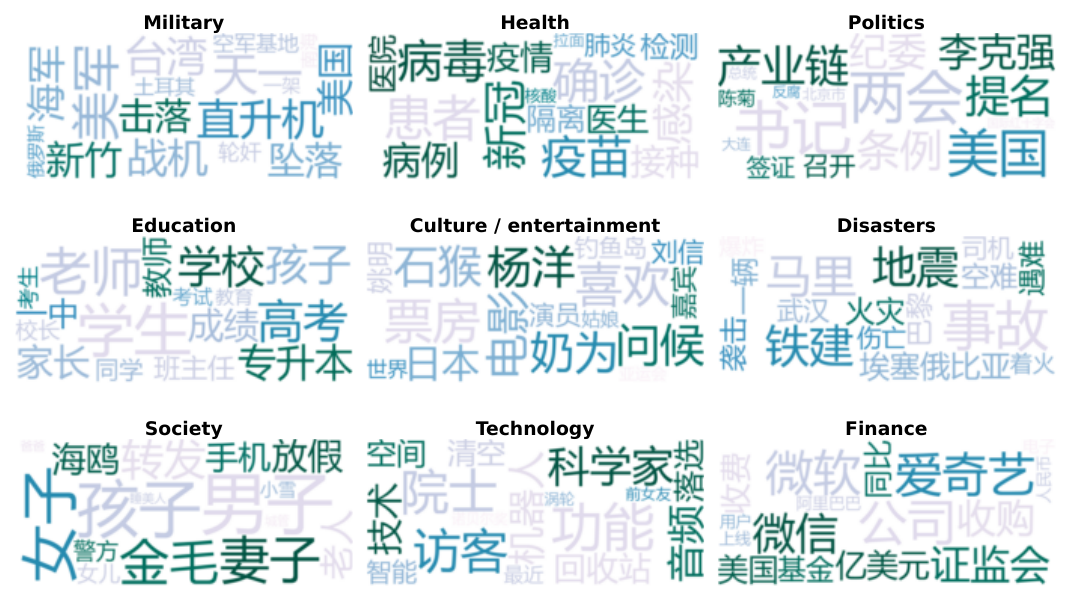}
    \caption{
        Word clouds for the nine named Weibo21 topics used in our evaluation.
        Larger terms occur more frequently within the corresponding topic.
    }
    \label{fig:weibo21-topic-wordclouds}
\end{figure*}

\section{Additional Weibo21 Details}
\subsection{Dataset Composition and Topic-Balanced Sampling}
\label{app:weibo21-sampling}

The official Weibo21 release~\cite{nan2021mdfend} contains 5,751 training, 1,918 validation, and 1,923 test cases, comprising 9,592 cases in total. It provides nine named topics together with an additional undetermined category. Table~\ref{tab:weibo21-composition} reports the distribution of these categories across the three official splits.

\begin{table}[h]
    \centering
    \footnotesize
    \setlength{\tabcolsep}{3.0pt}
    \renewcommand{\arraystretch}{1.12}
    \begin{tabular*}{\columnwidth}{@{\extracolsep{\fill}}lrrrrr@{}}
        \toprule
        \textbf{Topic}
        & \textbf{Train}
        & \textbf{Val.}
        & \textbf{Test}
        & \textbf{Total}
        & \textbf{Selected} \\
        \midrule
        Military      & 205   & 69  & 69  & 343   & 150 \\
        Health        & 600   & 200 & 200 & 1,000 & 150 \\
        Politics      & 511   & 170 & 171 & 852   & 150 \\
        Education     & 294   & 98  & 99  & 491   & 150 \\
        Culture/Ent.  & 864   & 288 & 288 & 1,440 & 150 \\
        Disasters     & 465   & 155 & 156 & 776   & 150 \\
        Society       & 1,601 & 534 & 534 & 2,669 & 150 \\
        Technology    & 141   & 47  & 48  & 236   & 150 \\
        Finance       & 792   & 264 & 265 & 1,321 & 150 \\
        Undetermined  & 278   & 93  & 93  & 464   & --  \\
        \midrule
        \textbf{Total}
        & \textbf{5,751}
        & \textbf{1,918}
        & \textbf{1,923}
        & \textbf{9,592}
        & \textbf{1,350} \\
        \bottomrule
    \end{tabular*}
    \caption{
        Topic distribution of the original Weibo21 dataset and the topic-balanced subset used in our evaluation.
        Val.\ denotes the official validation split.
    }
    \label{tab:weibo21-composition}
\end{table}

Because Weibo21 is used exclusively for evaluation, we pool its three official splits before constructing the topic-balanced subset. We exclude the undetermined category, remove duplicated news and duplicates with conflicting topic annotations. Using a fixed sampling seed of 2027, we select 75 real and 75 fake cases from each of the nine named topics. The resulting evaluation set contains 150 cases per topic and 1,350 cases in total. The same cases are used to construct the style, emotion, and dissemination views for all evaluated detectors.

\subsection{Detailed Zero-Shot Results}
\label{app:weibo21-zeroshot}
Table~\ref{tab:weibo21-zeroshot} reports the overall performance
of the four zero-shot LLMs on the topic-balanced Weibo21 set.
Unlike the topic-averaged results in
Appendix~\ref{sec:topic-analysis}, these results are computed
over all 1,350 cases without separating them by topic.

\begin{table}[t]
    \centering
    \footnotesize
    \renewcommand{\arraystretch}{1.15}

    \begin{tabular}{@{} l *{4}{r} @{}}
        \toprule
        \multicolumn{5}{@{}c}{\textbf{Panel A: Accuracy (\%)}} \\
        \addlinespace[3pt]
        \textbf{Model}
        & \textbf{Orig.} & \textbf{Style} & \textbf{Emo.} & \textbf{Dissem.} \\
        \midrule
        
        Qwen2.5-7B   & \textbf{77.04} & 57.85 & \textbf{60.37} & \textbf{55.78} \\
        Llama-3.1-8B & 66.96 & 49.11 & 43.19 & 36.00 \\
        Qwen3-8B     & 73.33 & \textbf{60.89} & 58.96 & 54.07 \\
        GLM-4-9B     & 65.48 & 55.33 & 50.74 & 50.52 \\
        \bottomrule
    \end{tabular}

    \vspace{1.5em}

    \begin{tabular}{@{} l *{5}{r} @{}}
        \toprule
        \multicolumn{6}{@{}c}{\textbf{Panel B: Macro-$F_1$ and UASR (\%)}} \\
        \addlinespace[3pt]
        \textbf{Model}
        & \textbf{Orig.} & \textbf{Style} & \textbf{Emo.} & \textbf{Dissem.} & \textbf{UASR~$\downarrow$} \\
        \midrule
        
        Qwen2.5-7B   & \textbf{76.87} & 57.61 & \textbf{60.36} & \textbf{55.49} & 53.08 \\
        Llama-3.1-8B & 66.34 & 43.15 & 41.08 & 32.95 & 62.61 \\
        Qwen3-8B     & 72.59 & \textbf{60.86} & 57.97 & 53.91 & 46.87 \\
        GLM-4-9B     & 62.68 & 47.37 & 44.49 & 41.50 & \textbf{37.67} \\
        \bottomrule
    \end{tabular}
    \caption{
        Overall zero-shot results on the topic-balanced Weibo21 evaluation set.
        All values are percentages.
        Higher Accuracy and Macro-F1 and lower UASR are better.
        Boldface denotes the best result in each column.
    }
    \label{tab:weibo21-zeroshot}
\end{table}
Qwen2.5-7B-Instruct achieves the strongest performance on the original, emotion, and dissemination views, whereas Qwen3-8B performs best under the style transformation. Nevertheless, all four models exhibit substantially lower Macro-F1 on at least one transformed view than on the original news. Llama-3.1-8B-Instruct is the most sensitive overall, with a UASR of 62.61\%. GLM-4-9B obtains the lowest UASR, but this result should be interpreted together with its comparatively low original-view and transformed-view performance because UASR conditions only on originally correct cases.

\subsection{Topic-Level Analysis}
\label{sec:topic-analysis}

We further investigate how news topics and presentation
transformations affect detection performance on Weibo21. Unlike
the comparison of English-trained detectors in
Section~\ref{sec:crosslingual}, this analysis averages the results
of four zero-shot LLMs and is intended to characterize topic- and
view-level performance variation.
Figure~\ref{fig:topic_performance_radial} presents Accuracy and
Macro-F1 across nine topics under the original view and the three
presentation transformations.

\label{app:weibo21-details}
Averaged across the nine topics, the zero-shot LLMs achieve a
Macro-F1 of 69.45\% on the original news, but their performance
decreases to 51.94\%, 50.76\%, and 45.71\% under the style,
emotion, and dissemination transformations, respectively.
Thus, their ability to classify unmodified Chinese news does
not translate into robustness to presentation changes.

Permutation tests based on 10,000 label-stratified case-level
reassignments indicate significant differences in overall
detection performance across topics for both Accuracy
($p_{\mathrm{perm}}=0.0061$) and Macro-F1
($p_{\mathrm{perm}}=0.0053$).
Consistent with Figure~\ref{fig:topic_performance_radial},
military news is generally easier to classify, whereas
education and disaster-related news are more difficult.
In contrast, we find no significant evidence that the relative
effects of the three transformations vary across topics:
the Topic~$\times$~Transformation interaction is not significant
for Accuracy ($p_{\mathrm{perm}}=0.0617$) or Macro-F1
($p_{\mathrm{perm}}=0.0781$).
Dissemination framing nevertheless yields the lowest Accuracy
and Macro-F1 among the three transformations in every topic,
making it the most consistently challenging transformation
in this evaluation.

\subsection{Topic Word Clouds}
\label{app:weibo21-wordclouds}

Figure~\ref{fig:weibo21-topic-wordclouds} provides a qualitative overview of the recurring terms associated with the nine Weibo21 topics included in our evaluation. The original three-by-three visualization is rearranged into a single-column layout to preserve the readability of the Chinese terms.

\end{document}